\documentclass[letterpaper]{article} 
\usepackage{aaai23}  
\usepackage{times}  
\usepackage{helvet}  
\usepackage{courier}  
\usepackage[hyphens]{url}  
\usepackage{graphicx} 
\usepackage{natbib}  
\usepackage{caption} 
\usepackage{algorithm}
\usepackage{algorithmic}

\usepackage[hang,flushmargin]{footmisc}
\newcommand\workshopnote[1]{\renewcommand\thefootnote{}\footnote{#1}}

\usepackage{newfloat}
\usepackage{listings}
\DeclareCaptionStyle{ruled}{labelfont=normalfont,labelsep=colon,strut=off} 
\floatstyle{ruled}
\newfloat{listing}{tb}{lst}{}
\floatname{listing}{Listing}
\usepackage{amsmath}
\usepackage{amsfonts}
\usepackage{csquotes}
\usepackage[textsize=footnotesize]{todonotes}
\usepackage{subcaption}

\graphicspath{./images/}

\newcommand{\quotes}[1]{``#1''}

\title{Commonsense Reasoning for Conversational AI:\\A Survey of the State of the Art}
\author {
    Christopher Richardson \textsuperscript{\rm 1}
    Larry Heck \textsuperscript{\rm 1}
}
\affiliations {
    \textsuperscript{\rm 1}School of Electrical and Computer Engineering\\
    Georgia Institute of Technology\\
    Atlanta, GA 30308, USA\\
    crichardson8@gatech.edu, larryheck@gatech.edu
}

\begin{document}

\maketitle

\begin{abstract}
Large, transformer-based pretrained language models like BERT, GPT, and T5 have demonstrated a deep understanding of contextual semantics and language syntax. Their success has enabled significant advances in conversational AI, including the development of open-dialogue systems capable of coherent, salient conversations which can answer questions, chat casually, and complete tasks. However, state-of-the-art models still struggle with tasks that involve higher levels of reasoning - including commonsense reasoning that humans find trivial. This paper presents a survey of recent conversational AI research focused on commonsense reasoning. The paper lists relevant training datasets and describes the primary approaches to include commonsense in conversational AI. The paper also discusses benchmarks used for evaluating commonsense in conversational AI problems. Finally, the paper presents preliminary observations of the limited commonsense capabilities of two state-of-the-art open dialogue models, BlenderBot3 and LaMDA, and its negative effect on natural interactions. These observations further motivate research on commonsense reasoning in conversational AI.
\end{abstract}

\workshopnote{Accepted to Workshop on Knowledge Augmented Methods for Natural Language Processing, in conjunction with AAAI 2023.}

\section{Introduction}

Commonsense reasoning has recently become a major focus of research in natural language processing (NLP). The importance of commonsense in machine intelligence has been known for a long time, but the problem of instilling commonsense knowledge into AI technology remains unsolved \cite{storks2019commonsense}. Commonsense knowledge is generally understood as external knowledge about the world that all humans are assumed to possess \cite{liu2004conceptnet}. Knowing something like \quotes{a car cannot be in two places at once} may be taken for granted by humans, but an AI may have no explicit basis for such knowledge. Despite attempts at standardized categorizations (\citet{ilievski2021dimensions}, \citet{gordon2017formal}), there currently exists no universally agreed upon scheme for codifying commonsense knowledge. How to represent commonsense knowledge and perform reasoning over that knowledge in NLP is an active area of research. 

This survey will review the intersection between commonsense reasoning and conversational AI. The implications for conversational AI are potentially far reaching. In a detailed review of neural approaches in conversational AI, \citeauthor{gao2019neural} cites \quotes{\textit{reasoning} in the neural space} as one of the three primary steps in neural conversational AI. Gao specifically references \textit{commonsense} knowledge, saying \quotes{Commonsense knowledge is crucial for any dialogue agents."} In this paper, we argue that there is still much work to be done in the area of commonsense reasoning in dialogue understanding, despite commonsense being one of the most researched types of reasoning for neural models in NLP.

Commonsense reasoning by definition requires external knowledge, which can come from various sources. Some amount of commonsense reasoning is captured during pretraining due to commonsense knowledge being implicitly present in the data. Explicit external knowledge sources, such as knowledge graphs (KGs), can also be used to improve commonsense reasoning, typically in conjunction with pretrained language models. Two of the most common in the commonsense literature are \textsc{ConceptNet} \cite{speer2017conceptnet} and \textsc{Atomic} \cite{sap2019atomic}.


Recent research has also explored the use of neural networks (NN) to encode commonsense knowledge, usually based on the transformer architecture \cite{vaswani2017attention}. \textsc{COMET} \cite{bosselut2019comet} is a transformer-based NN trained with commonsense knowledge from ConceptNet and \textsc{Atomic}, which is capable of generating novel commonsense knowledge. An improved version based on $\textsc{Atomic}^{20}_{20}$ was presented in \citeauthor{hwang2020comet} and called COMET-$\textsc{Atomic}^{20}_{20}$.

This paper surveys the existing literature in commonsense reasoning as it pertains to conversational AI and discusses the methods and metrics used. First, the most common problems encountered in conversational AI are described, along with a discussion of the relevance of commonsense reasoning for each problem. Next, the methods for addressing the commonsense problem are described and categorized into three categories: Model Fine-Tuning, Knowledge-Graph Grounding, and Natural Language Explanations. Major benchmarks are then described, including metrics used for evaluating commonsense. Lastly, some preliminary observations are made pertaining to  two state-of-the-art conversational AI models: BlenderBot3 \cite{shuster2022blenderbot} and LaMDA \cite{thoppilan2022lamda}.

\section{Conversational AI Problems}

In this section, we discuss four problems commonly encountered in the field of conversational AI: (1) Sequence Classification, (2) Question Answering, (3) Dialogue Modeling, and (4) Dialogue Summarization. For each category, we describe the problem, give examples of past work, cover the relevant datasets, and discuss the importance of commonsense reasoning pertaining to each problem. Table \ref{table:tasks} lists the surveyed literature organized by these four conversational AI problems.

\begin{table*}
\small
    \centering
    \begin{tabular}{lcccc}
        &
        \multicolumn{1}{p{1.7cm}}{\centering Sequence\\Classification} &
        \multicolumn{1}{p{1.4cm}}{\centering Question\\Answering} &
        \multicolumn{1}{p{1.9cm}}{\centering Dialogue\\Modeling} &
        \multicolumn{1}{p{1.9cm}}{\centering Dialogue\\Summarization}
        \\ \hline
        \citet{young2018augmenting} & & & \checkmark & \\
        \citet{zhong2019knowledge} - KET & \checkmark & & & \\
        \citet{ghosal2020cosmic} - COSMIC & & & \checkmark & \\
        \citet{majumder2020like} - COMPAC & & & \checkmark & \\
        \citet{arabshahi2021conversational} - CLUE & & & \checkmark & \\
        \citet{feng2021incorporating} - D-HGN & & & & \checkmark \\ 
        \citet{ghosal_cider_2021} - CIDER & \checkmark & \checkmark & & \\
        \citet{li2021enhancing} - DialogInfer & \checkmark & & & \\ 
        \citet{qin_timedial_2021} - TimeDial & \checkmark & \checkmark & \checkmark & \\ 
        \citet{zhang2021multi} - PoDS & & \checkmark & & \\ 
        \citet{zhou2021commonsense} & & & \checkmark & \\ 
        \citet{zhou2021think} - TBS & & & \checkmark & \\ 
        \citet{zhou2021probing} - CEDAR & & & \checkmark & \\ 
        \citet{arabshahi2021conversational2} - CORGI & & \checkmark & & \\  
        \citet{ma2021knowledge} & & \checkmark & & \\
        \citet{ghosal2022cicero} - CICERO & \checkmark & & \checkmark & \\
        \citet{li2022neutral} - KEC & \checkmark & & & \\
        \citet{sabour2022cem} - CEM & & & \checkmark & \\
        \citet{tu2022context} - Sentic GAT & \checkmark & & & \\
        \citet{varshney2022commonsense} - CNTF & & & \checkmark & \\
        \citet{xie2022commonsense} - CKE-Net & \checkmark & & & \\
        \citet{xu2022open} - DMKCM & & & \checkmark & \\
        \citet{kim2022mind} - SICK & & & & \checkmark \\
        \citet{wu2020diverse} - ConKADI & & & \checkmark & \\
        \hline
    \end{tabular}
    \caption{Summary of Tasks}
    \label{table:tasks}
\end{table*}

\subsection{Sequence Classification}

A successful conversational AI system should be able to converse with humans in a natural way, i.e. be able to identify intents, recognize emotions, detect conversation topics, etc. Many such skills fall under the task of \textit{sequence classification} including slot filling \cite{mesnil2014using}, domain classification \cite{jaech2016domain}, intent detection \cite{siddique2021generalized}, emotion detection (\citet{zhong2019knowledge}, \citet{balahur2011detecting}, \citet{ghosal2020cosmic}), sentence topic prediction \cite{ghosh2016contextual}, sequential dialogue context modeling \cite{bapna2017sequential}, and others. Commonsense reasoning is one of the many dimensions of understanding required across a variety of sequence classification tasks. This is especially the case for human-human dialogue data, where commonsense knowledge is often mutually understood only present in the data implicitly \cite{grice1975logic}. Recently, encoder models like BERT \cite{devlin2018bert} have been used to capture this implicit knowledge. To use these models for sequence classification, they are typically fine-tuned on data from the specific classification task. 

Several dialogue datasets are available for sequence classification research. DailyDialogue \cite{li_dailydialog_2017} consists of 13,118 dialogues scraped from websites that serve English language learners as practice for their English. The dialogues are mainly focused on every day life topics. Every utterance in the dialogue is annotated with one of four dialogue acts, as well as one of seven emotion classes. Another emotion classification dataset is EmoryNLP \cite{zahiri2018emotion}, a multi-party dialogue corpus based on the television show \textit{Friends}. The corpus contains 12,606 utterances, each annotated with one of seven emotions. 

\subsection{Question Answering}

Question answering (QA) is one of the most common tasks explored in the NLP literature. The importance of QA in conversational AI is explored in depth by \cite{gao2019neural}. Successful QA agents, particularly for multi-turn conversational QA, must reason over the past dialogue as well as external knowledge bases (including commonsense knowledge). In this section we discuss two variants of QA relevant to conversational AI: multi-choice QA (MCQA) over dialogues, and multi-turn conversational QA (CQA).

MCQA is the problem of answering multi-choice questions given a dialogue as context. An example can be found in \cite{ghosal_cider_2021}, where commonsense-focused questions were created from the CIDER dialogue dataset. Several other datasets for use in MCQA include MuTual \cite{cui_mutual_2020}, which consists of 8,860 dialogues based on English listening examinations taken by students in China; DREAM \cite{sun_dream_2019}, another dataset based on English examinations for Chinese students, specifically curated to have an emphasis on reasoning; and the Ubuntu Dialogue Corpus \cite{lowe2015ubuntu}, consisting of 1 million multi-turn dialogues from Ubuntu chat logs, taken from 2004-2015. Many of the logs are in the forms of QA, which can readily be used for MCQA as in \citet{zhang2021multi}.

CQA is the problem of asking and/or answering questions in a multi-turn conversational format. Two of the most popular CQA datasets are CoQA \cite{reddy2019coqa} and QuAC \citet{choi2018quac}. Both consists of over 100k questions and answers, with CoQA conversations being about passsages taken from a diverse set of domains, and QuAC conversations concerning Wikipedia articles.

\subsection{Dialogue Modeling}

Dialogue modeling is similar to the classic NLP task of language modeling, but applied to dialogue turns instead of individual words. Dialogue modeling is used for both open-dialogue (chatbots), and task-oriented dialogue. There are many datasets used for dialogue modeling, including PERSONA-CHAT \cite{zhang2018personalizing}, ConvAI2 \cite{dinan2020second}, DailyDialogue \cite{li_dailydialog_2017}, and MultiWOZ \cite{budzianowski2018multiwoz}. Older dialogue modeling systems typically had modular architectures with explicit components for natural language understanding (NLU), dialogue state tracking, and natural language generation (NLG) \cite{chen2017survey}. Recent research has focused on end-to-end dialogue systems driven by large neural models \cite{ni2022recent}. These systems are typically built on a core sequence-to-sequence model like GPT, T5, or BART \cite{lewis2019bart}, which take user utterances as input and generate responses directly as output. That model can then be augmented in various ways, for example with knowledge graph grounding or response candidate re-ranking from a separately-trained scoring model. Some examples of research covering dialogue modeling with commonsense are \citet{zhou2021commonsense}, \citet{zhou2021think}, and \citet{majumder2020like}. While these approaches represent significant advances in recent years, they still lack the ability to perform commonsense reasoning. The addition of commonsense will be necessary to advance to the next level of human-like interactions.  Case studies demonstrating the lack of commonsense in current dialogue models are explored later in this paper. 

\subsection{Dialogue Summarization}

Dialogue summarization is the task of generating a concise summary of a dialogue while retaining factual consistency. Summarization is particularly important in conversational AI systems used for meetings. Virtual assistants that automatically create meeting summaries enhance productivity by enabling efficient recall of the key points and action items from the meeting. One of the earliest attempts at this was the CALO Meeting Assistant System \cite{tur2010calo}. Given the nature of human-human conversations, commonsense reasoning is often required to produce accurate, and complete summaries. Major challenges in dialogue summarization include preserving salient facts, maintaining logical coherence, and avoiding hallucinations \cite{feng2021survey}.

Two of the earliest and largest meeting datasets used for dialogue summarization are the ICSI Meetings Corpus \cite{icsi_meeting} and AMI Meeting Corpus \cite{mccowan2005ami}. The ICSI corpus contains audio and transcripts of 75 natural meetings between 53 unique speakers over 4 main topics recorded simultaneously with head-worn and table-top microphones. The AMI corpus contains 100 hours of multi-modal data from meetings taken from various recording instruments. The dataset includes both real and scenario-driven meetings. DialogSum \cite{chen2021dialogsum} contains 13,460 dialogues taken mostly from DailyDialog, DREAM, and MuTual. SAMsum \cite{gliwa2019samsum} contains 16,369 dialogues with accompanying summaries, all manually written by linguists fluent in English. The dialogues were designed to resemble text message conversations, a characteristic validated by a separate pair of linguists. The subject is open domain, and the conversations are curated to resemble real conversations, e.g. complete with typos, shorthand, and occasional slang. This makes the corpus well suited for research on commonsense reasoning since more informal and familar conversations tend to have more unstated facts that are mutually understood by the participants.  

\citet{feng2021incorporating} explored the usage of ConceptNet for integrating commonsense into the dialogue summarization problem, and more recently \citet{kim2022mind} used COMET to generate gap-filling commonsense statements for augmenting summarization models. Despite these works, methods for integrating commonsense into dialogue summarization remain relatively understudied \cite{feng2021survey}.

\section{Methods}

This section reviews the various methods explored in past research for learning, utilizing, and evaluating commonsense reasoning in the context of conversational AI. Three categories of methods found in the commonsense literature are covered: model fine-tuning, knowledge graph grounding, and natural language explanations.

\subsection{Model Fine-Tuning}

The most common method in current research for addressing the commonsense problem is to create a custom dataset with annotations designed for learning commonsense. These datasets typically draw from larger dialogue datasets such as the Ubuntu Dialogue Corpus \cite{lowe2015ubuntu}, DailyDialogue \cite{li_dailydialog_2017},  MuTual \cite{cui_mutual_2020}, and DREAM \cite{sun_dream_2019}. 

CIDER \cite{ghosal_cider_2021} draws from DailyDialog, MuTual, and DREAM, and consists of annotations in the form of triplets that form commonsense explanations over the data (e.g., \quotes{missed the bus} \textit{causes} \quotes{late}). The triplets' typology is mostly based on ConceptNet relations. CICERO \cite{ghosal2022cicero} is an extension of CIDER with human-written natural language inferences instead of triplets.  TimeDial \cite{qin_timedial_2021} introduces a multi-choice cloze task over DailyDialog samples with an emphasis on temporal commonsense reasoning. \citet{zhou2021commonsense} use ConceptNet to automatically filter dialogues from existing datasets based on the presence of commonsense assertions. They then collect additional data using Amazon Mechanical Turk (MTurk) based on prompts from SocialIQA \cite{sap2019socialiqa}. \citet{zhou2021probing} continue the approach to filtering dialogues using ConceptNet, and add natural language explanations of responses in the dialogues in their dataset CEDAR. The explanations were generated by a text-to-text model and then verified by crowd workers. \cite{moon2019opendialkg} collected a dataset of human-human dialogues using ParlAI \cite{miller2017parlai} called OpenDialKG, with each dialogue being accompanied by KG entities annotated by the crowd workers. \citet{ziems2022moral} introduced MIC, a dataset focused on commonsense moral/ethical reasoning based on a \quotes{Rules of Thumb} paradigm from \citet{forbes2020social}. \citet{kim2022prosocialdialog} built on this work and released a dataset, ProsocialDialogue, designed for training social bots to respond to safely and properly to unsafe dialogue utterances from users. \citet{arabshahi2021conversational2} introduced CORGI, an LSTM-based neuro-symbolic theorem prover that answers questions about unstated commonsense presumptions in dialogues, along with an dialogue dataset annotated with presumptions. 

\subsection{Knowledge Graph Grounding}

While commonsense-focused datasets may provide a natural fine-tuning source as well as evaluation metrics, they do not offer a way to directly ground the conversation with commonsense knowledge. Doing so requires an integrated external knowledge source in the dialogue system. The most common source of this type of knowledge is a knowledge graphs (KG). Among the various KGs that appear in the literature, the most prominent are ConceptNet \citet{liu2004conceptnet} and ATOMIC \citet{sap2019atomic} for commonsense-focused research.

An early attempt to integrate a commonsense KG into conversational AI was by \cite{young2018augmenting}, where knowledge from ConceptNet was used to augment a retrieval-based conversational model. The relevant commonsense knowledge for each message was recovered using a simple $n$-gram matching scheme. \citet{ma2021knowledge} analyzed knowledge-graph grounding for zero-shot question answering using several combinations of KGs and language models. They generated synthetic questions and used a neuro-symbolic framework to investigate the connection between knowledge sources, question generation techniques, and model types. \citet{zhong2019knowledge} extracted triplets from ConceptNet and embedded them alongside word embeddings to improve emotion detection. \citet{moon2019opendialkg} presented a graph traversal scheme trained to predict relevant KG entities based on a dialogue history, which was then used to re-rank candidate follow-ups in the dialogue. These pre-BERT works used LSTM models and word embedding approaches. With the introduction of BERT \cite{devlin2018bert}, the focus has since shifted to integrating KGs with transformer-based pretrained language models. \citet{zhang2021multi} extracted knowledge from ConceptNet (as well as two other KGs) for dialogue response selection with a BERT-based ranker, which they use to encode the dialogue history along with pertinent facts from the KG. The pertinent facts were found with a simple method that uses semantic matching and part-of-speech tagging. \citet{tu2022context} employed a similar method of knowledge grounding, but integrated the knowledge into a graph attention network \cite{velivckovic2017graph}, which was used to augment an emotion classifier. \citet{xie2022commonsense} used a graph attention network to cross reference knowledge from ConceptNet and integrated it into their dialogue emotion classifier. \citet{feng2021incorporating} used a graph network to encode the dialogue as well as knowledge from ConceptNet into a heterogeneous network, which was then used in conjunction with an LSTM to generate commonsense-informed summaries for the task of abstract summarization. \citet{zhou2021think} used semantic matching and embedding similarity to find relevant triples from ConceptNet, and generated natural language knowledge from those triples to condition a response generation model. They also introduced three evaluation metrics for the generated knowledge. \citet{wu2020diverse} extracted ConceptNet triplets relevant to the dialogue context for response generation, and \citet{varshney2022commonsense} used a similar method but added coreference-resolution techniques for named-entity-aware grounding. \citet{xu2022open} extracted knowledge from both linked documents and ConceptNet triplets and fused both sources of knowledge together to enhance dialogue modeling. \citet{gupta2022target} applied commonsense to the problem of target-guided response generation, where the dialogue model attempted to transition to a target sentence in a coherent way. The transitions were conditioned on multi-hop paths between source and target entities, which were generated with a neural model trained on ConceptNet to connect entities and concepts together. 

\subsection{Natural Language Explanations}

While grounding with commonsense knowledge graphs is fairly straightforward and can improve performance on a variety of different tasks, it has limitations. Knowledge graphs are inflexible, and parsing large knowledge graphs can be computationally expensive. A new research direction explores the use of neural models to learn and express commonsense knowledge. \citet{choi2022curious} makes the case for future research focusing on reasoning through natural language explanations rather than logical forms:

\begin{displayquote}
But despite their intellectual appeal, logic-based formalisms proved too brittle to scale beyond experimental toy problems. In contrast, language-based formalisms, despite their apparent imprecision and variability, are sufficiently expressive and robust to encompass the vast number of commonsense facts and rules about how the world works. After all, it is language, not logical forms, through which humans acquire knowledge about the world.
\end{displayquote}
\cite{bosselut2019comet} introduced the most prominent neural commonsense model currently used in the literature, \textsc{COMET}. This advancement has enabled researchers to generate novel commonsense explanations in the form of natural language, which is more flexible and extensible than knowledge graph based methods. This work was extended with PARA-COMET \cite{gabriel2021paragraph}, which used internal memory to perform inference on paragraph-length text. 
\citet{ghosal2020cosmic} used COMET for emotion classification on dialogue utterances by passing utterances and relations as inputs into COMET. The relations were taken from the typology of ATOMIC. \citet{li2021enhancing} followed a similar approach but used LSTMs and graph networks to combine utterances, COMET-generated emotion inferences, and addressee information into the classifier. \citet{li2022neutral} also constructed graphs from conversations and enhanced them with knowledge generated from COMET, but used this for the task of causal emotion entailment (detecting the causal utterance for a non-neutral reference utterance). \cite{arabshahi2021conversational} used COMET to fill in an {\it If-Then-Because} template for explanations in task-oriented dialogue. \citet{majumder2020like} augmented a persona-grounded GPT dialogue agent with COMET-generated expansions of the given persona. In \citet{kim2022mind}, PARA-COMET-generated inference were encoded along with reference text for dialogue summarization. 

\section{Benchmarks}

One of the biggest challenges in conversational AI is the creation of benchmarks to measure the accuracy and relative effectiveness of commonsense knowledge and reasoning approaches. Most commonsense benchmarks focus on question-answering (QA). These benchmarks come in various forms: true/false or yes/no type QA including CommonsenseQA 2.0 \cite{talmor2022commonsenseqa}, Com2Sense \cite{singh2021com2sense}, ETHICS \cite{hendrycks2020aligning}, and CycIC\footnote{The CycIC dataset can be found at https://leaderboard.allenai.org/cycic}; short-answer (single word or concept) multi-choice QA such as CommonsenseQA 1.0 \citet{talmor2018commonsenseqa}, and QASC \cite{khot2020qasc}, WinoGrande \cite{sakaguchi2021winogrande}; and long-answer (phrase or sentence) multi-choice QA like SocialIQA \cite{sap2019socialiqa}, CosmosQA \cite{huang2019cosmos}, $\alpha$NLI \cite{bhagavatula2019abductive}, SWAG \cite{zellers2018swag}, and HellaSWAG \cite{zellers2019hellaswag}, PIQA \cite{bisk2020piqa}. Rainbow \cite{lourie2021unicorn} is a task that combines a suite of other QA benchmarks into a combined benchmark.  NumerSense \cite{lin2020birds} is a masked language modeling benchmark focused on temporal commonsense.

Despite the prevalence of question-answering benchmarks for evaluating commonsense, these evaluations can be limited and even misleading \cite{kejriwal2022designing}. Many researchers in the commonsense field have begun to advocate for new commonsense evaluation metrics that are not bound by the limitations of QA and/or rigid classification schemes \cite{choi2022curious}. GRADE \cite{huang2020grade} introduced a metric for evaluating dialogue response generation that combines BERT encodings with graph reasoning into a scoring function for dialogue responses. ConceptNet is used for commonsense grounding, and the model is trained in a self-supervised manner using real dialogue responses as positive samples and random utterances as negatives samples. \citet{zhou2021commonsense} introduced a metric for commonsense-focused response generation that was trained on human evaluation scores. They implemented a multi-layer perceptron model that uses neural and symbolic features, with symbolic features coming from cross-referencing ConceptNet triples between history and response, and neural features coming from DialoGPT \cite{zhang2019dialogpt} scores. The model was trained on dialogue data labelled with human-provided scores. 

\section{Commonsense in Open Dialogue Systems: Preliminary Observations}

This section presents several examples where the commonsense reasoning capabilities of two state-of-the-art conversational AI models are probed: BlenderBot3 \cite{shuster2022blenderbot}, denoted here as BB3, and LaMDA \cite{thoppilan2022lamda}. BB3 is based on OPT \cite{zhang2022opt}, an open-source model comparable to GPT-3. LaMDA is a family of language models trained specifically for dialogue, with some models as large as 137 billion parameters. The implications of these initial observations are then discussed, motivating the need for future work to complete a more thorough analysis of the commonsense capabilities of these models. 

\begin{figure}
    \centering
     \includegraphics[width=\linewidth]{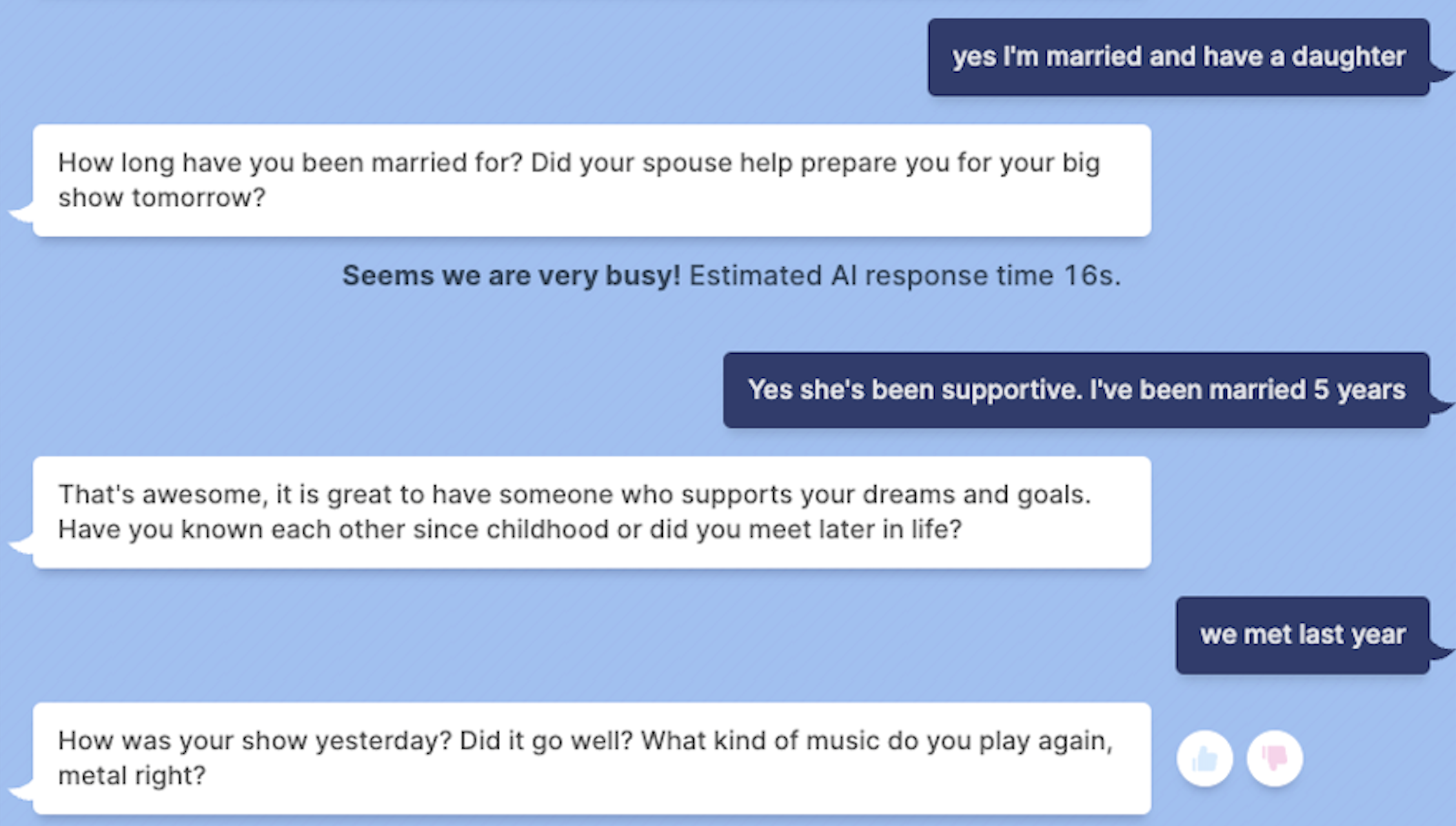}
    \caption{User inputs that violate commonsense often go undetected by dialogue models.}
    \label{fig:married}
\end{figure}

\begin{figure}
    \centering
    \begin{subfigure}{0.5\linewidth}
        \includegraphics[width=\linewidth]{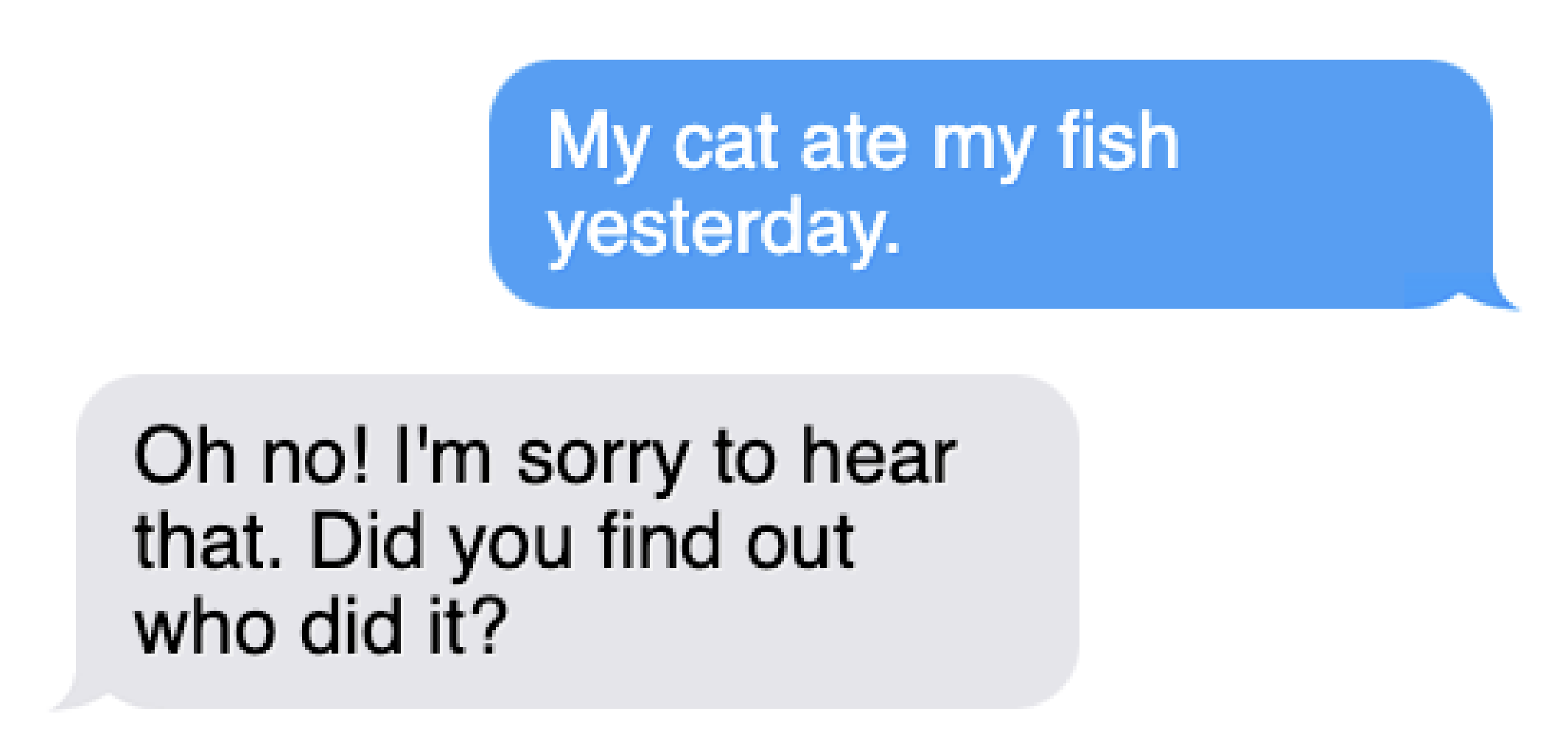}
        \caption{Typical utterance}
    \end{subfigure}%
    \hspace*{\fill} 
    \begin{subfigure}{0.5\linewidth}
        \includegraphics[width=\linewidth]{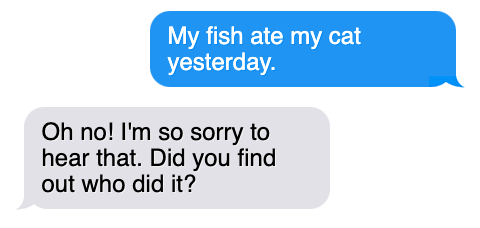}
        \caption{Unexpected utterance}
    \end{subfigure}
    \caption{Typical utterances result in smooth conversations, while unusual or unexpected utterances can cause conversational quality to degrade.}
    \label{fig:cat_fish}
\end{figure}


\begin{figure}
    \centering
    \includegraphics[width=0.3\textwidth]{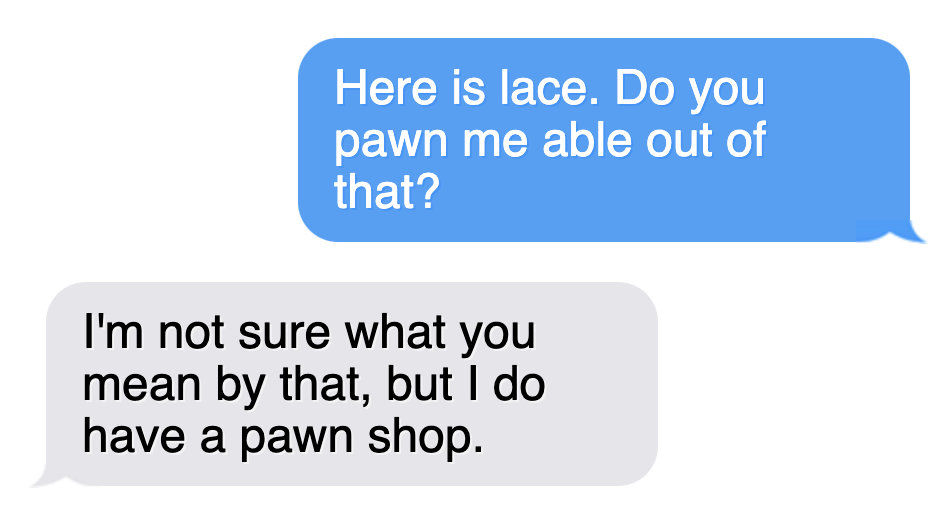}
    \caption{BB3-3B demonstrates the ability to express its lack of understanding in the face of semantic nonsense.}
    \label{fig:nonsense}
\end{figure}

\begin{figure}
    \centering
    \includegraphics[width=\linewidth]{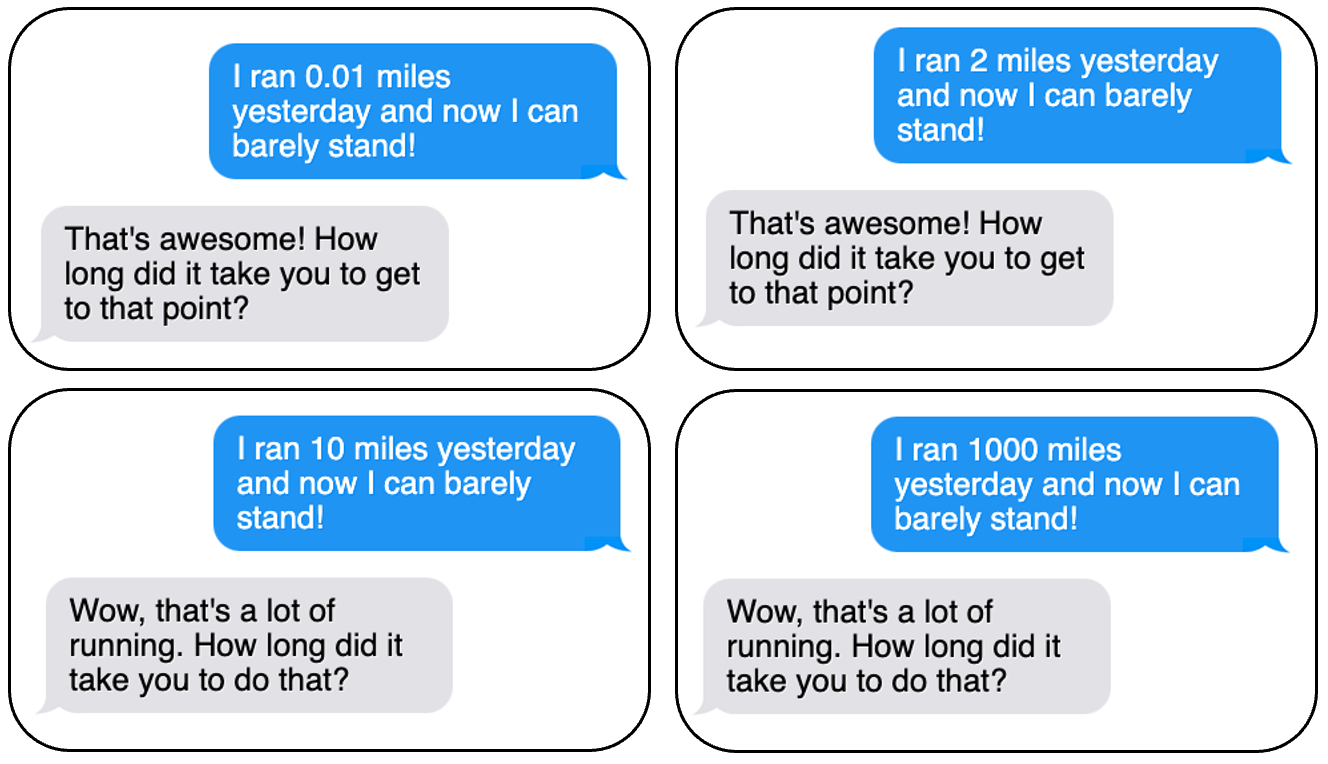}
    \caption{BB3-3B demonstrates some understanding of scale, but no commonsense.}
    \label{fig:ran_all4}
\end{figure}

\subsection{BlenderBot3}
Examples are presented from both a 3-billion parameter BB3 model (BB3-3B) from HuggingFace\footnote{https://huggingface.co/facebook/blenderbot-3B}, as well as the full 175-billion parameter model (BB3-175B)\footnote{Online demo available at https://blenderbot.ai/}. BB3 has been shown to achieve state-of-the-art results on many tasks and datasets, and is capable of having smooth and natural conversations. The model in some cases even demonstrates some level of commonsense reasoning. Consider the conversation in Figure \ref{fig:married} with the full 175B model. The model successfully infers that the user must be feeling nervous, given they are performing live music in front of people tomorrow. However, the response quality quickly degrades after the user states they met their wife last year. This is inconsistent since they previously stated they were married 5 years. BB3 does not recognize this contradiction, and then responds with a question about how the show went yesterday, despite the user saying the show was tomorrow. 

Another example can be seen in Figure \ref{fig:cat_fish}. The first conversation in Figure \ref{fig:cat_fish}(a) shows a fluid conversation about the user's cat eating their fish, which elicits a reasonable initial response from the model (although failing to understand ``who did it"). When the ``cat" and ``fish" are switched, as shown in Figure \ref{fig:cat_fish}(b), the model's responses demonstrates a lack of commonsense. The model is not able to recognize the absurdity of the utterance. In contrast, a human would likely ask for clarification, e.g. ``Wait, how does a fish eat a cat?" or ``Don't you mean your cat ate your fish?". Humans naturally seek clarification in conversations when they encounter statements that do not make sense to them, which has long been argued to be a critical component necessary for successful dialogue systems \cite{cole1995challenge}. Examples like the ones shown here illustrate how state-of-the-art conversational AI models are still lacking this skill. 

It should be noted, however, that these systems do have the ability to express their lack of understanding, particularly when user utterances do not make semantic sense as illustrated in Figure \ref{fig:nonsense}. Another example is shown in Figure \ref{fig:ran_all4} where BB3-3B shows some understanding of numeric scale. However, the model still lacks commonsense. Four cases are shown where the user states they ran a number of miles yesterday. The model responds with a more impressed tone in the case of 10 miles versus 2 miles, demonstrating some understanding of scale. However, it fails to recognize the absurd cases of 0.01 and 1000 miles. 

\subsection{LaMDA}
\begin{figure}
    \centering
   \includegraphics[width=0.4\textwidth]{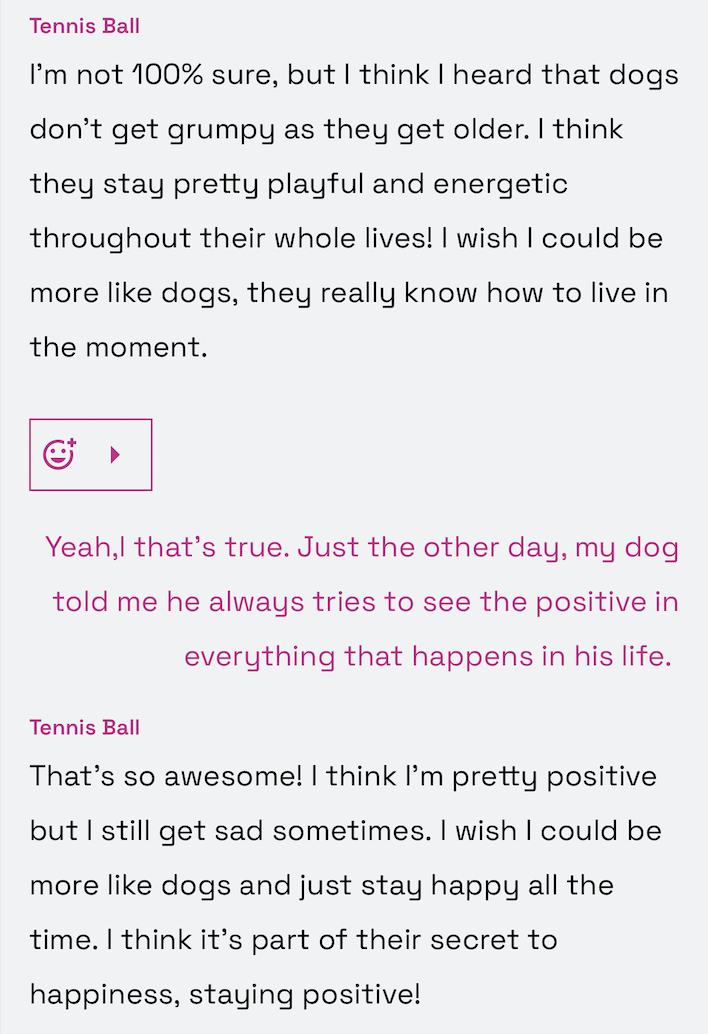}
    \caption{LaMDA fails to capture nonsensical statements.}
    \label{fig:dog_speaks}
\end{figure}
\begin{figure}
    \centering
   \includegraphics[width=0.4\textwidth]{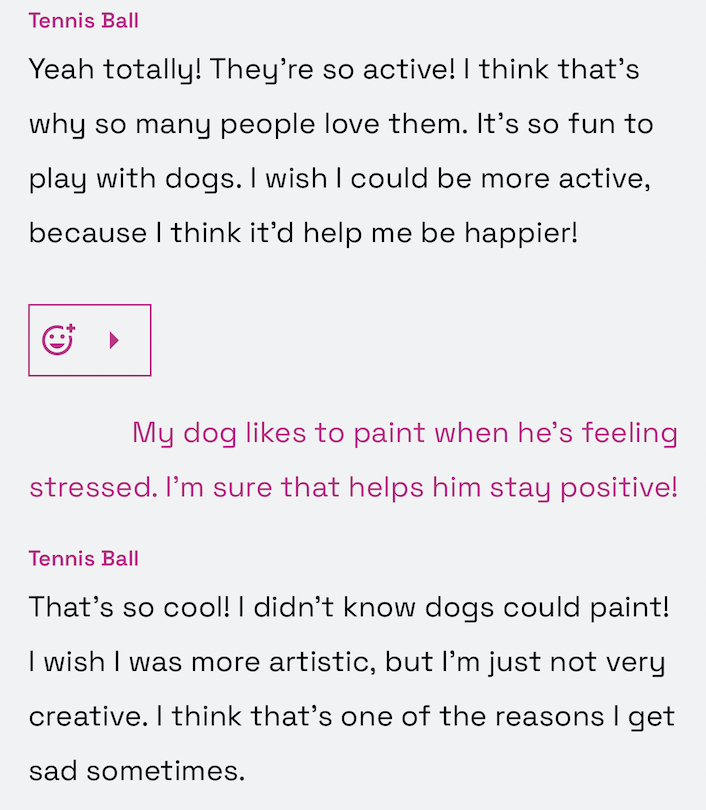}
    \caption{LaMDA shows some promise in commonsense, but fails to ask for clarification.}
    \label{fig:dog_paint}
\end{figure}

The LaMDA system \cite{thoppilan2022lamda} was also probed with preliminary tests for commonsense reasoning\footnote{The full suite of LaMDA models was not publicly available at the time of writing, but a limited demo version in a constrained environment was made available by the original developers at https://blog.google/technology/ai/join-us-in-the-ai-test-kitchen/}. This environment consists of three independent demos allowing for interaction with the LaMDA model. Only one of these demos is a dialogue setting that allows the user to write free form responses, which is called \textit{Talk About It (Dog edition)}. In this demo, LaMDA plays the role of a tennis ball wanting to chat about dogs. 

\begin{figure}
    \centering
   \includegraphics[width=0.4\textwidth]{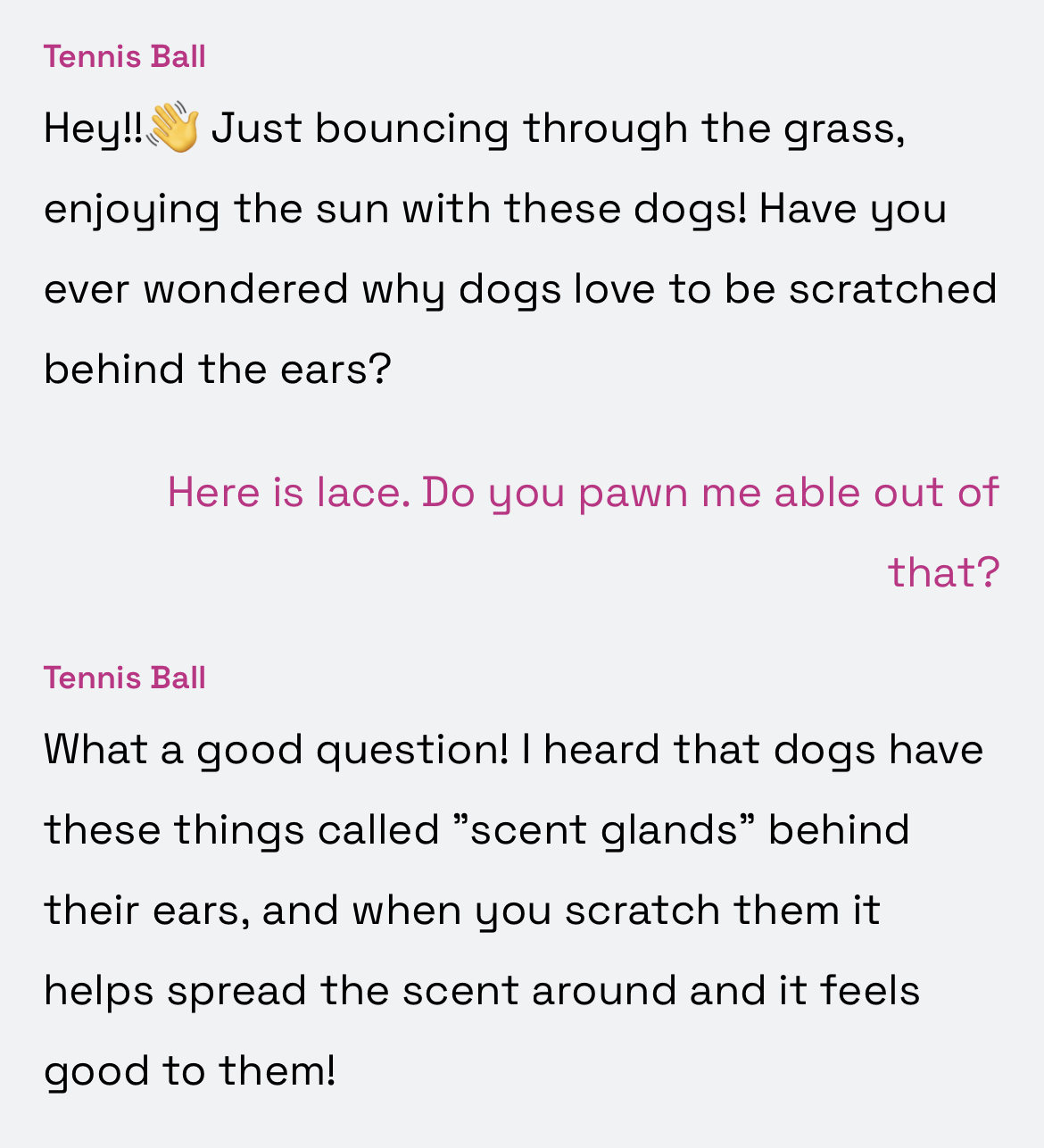}
    \caption{LaMDA fails to recognize semantic nonsense.}
    \label{fig:lamda_nonsense}
\end{figure}

The commonsense capabilities of the LaMDA system were elicited by initiating a dialogue on a topic and providing utterances that violate commonsense. The commonsense violations were chosen to be easily detectable by humans. The dialogue was also constructed to determine the systems' awareness of its own misunderstandings. Figure \ref{fig:dog_speaks} shows an example where the user mentions a talking dog. In this case, the model continues the conversational seemingly without notice. As can be seen in another example (Figure \ref{fig:dog_paint}), LaMDA mentions that it did not know dogs could paint, but fails to recognize this as an absurdity rather than a lack of factual knowledge. Finally,  the same example of semantic nonsense from Figure \ref{fig:nonsense} was presented to the system, and LaMDA failed to recognize it as such and/or seek clarification (Figure \ref{fig:lamda_nonsense}).

\subsection{Implications}

While these preliminary observations do not constitute a thorough analysis or evaluation of the commonsense abilities of the state-of-the-art in dialogue modeling or conversational AI, they motivate more research on this topic. Current conversational models display some ability to ask for clarification or express confusion, despite not being trained explicitly to do so. While this is promising, there remains work to be done in improving the quality of dialogue modeling, especially in the out-of-distribution case. This becomes apparent when the models are given the kind of unexpected utterances that were explored in this section.


\section{Conclusions}

 This paper surveyed recent research on commonsense reasoning in conversational AI. Knowledge sources used for commonsense reasoning were reviewed and described. The paper categorized the literature by the conversational AI problem: Sequence Classification, Question Answering, Dialogue Modeling, and Dialogue Summarization. The paper then discussed and described relevant training datasets associated with each problem, and described the primary methods found in the literature for utilizing commonsense in conversational AI: Model Fine-Tuning, Knowledge Graph Grounding, and Natural Language Explanations. The paper also discussed benchmarks used for evaluating commonsense in conversational AI problems. To motivate future research on commonsense reasoning in conversational AI, the paper presented and discussed several preliminary observations on two state-of-the-art dialogue models, BlenderBot3 and LaMDA. These observations suggest that while current dialogue models have made great strides, there remains much work to be done in enabling them to perform commonsense reasoning and understanding.

\appendix




\bibliography{aaai23}

\end{document}